\documentclass[11pt]{article}

\usepackage[preprint]{acl}

\usepackage{times}
\usepackage{latexsym}

\usepackage[T1]{fontenc}

\usepackage[utf8]{inputenc}

\usepackage{microtype}

\usepackage{inconsolata}

\usepackage{graphicx}

\usepackage{amsmath,amssymb,amsfonts}
\usepackage{multirow}
\usepackage{multicol}
\usepackage{makecell}

\title{AutoMedic: An Automated Evaluation Framework for Clinical Conversational Agents with Medical Dataset Grounding}

\author{
    \textbf{Gyutaek Oh\textsuperscript{1,2}},
    \textbf{Sangjoon Park\textsuperscript{*,2,3}},
    \textbf{Byung-Hoon Kim\textsuperscript{*,1,2,4,5}}
    \\
    \\
    \textsuperscript{1}Department of Biomedical Systems Informatics, Yonsei University College of Medicine,
    \\
    \textsuperscript{2}Yonsei Institute for Digital Health, Yonsei University
    \\
    \textsuperscript{3}Department of Radiation Oncology, Yonsei University College of Medicine,
    \\
    \textsuperscript{4}Department of Psychiatry, Yonsei University College of Medicine,
    \\
    \textsuperscript{5}Institute of Behavioral Sciences in Medicine, Yonsei University College of Medicine
    \\
    \small{
    \textbf{Correspondence:} \href{mailto:depecher@yuhs.ac}{depecher@yuhs.ac}, \href{mailto:egyptdj@yonsei.ac.kr}{egyptdj@yonsei.ac.kr}
    }
}

\begin{document}
\maketitle
\begin{abstract}
Evaluating large language models (LLMs) has recently emerged as a critical issue for safe and trustworthy application of LLMs in the medical domain.
Although a variety of static medical question-answering (QA) benchmarks have been proposed, many aspects remain underexplored, such as the effectiveness of LLMs in generating responses in dynamic, interactive clinical multi-turn conversation situations and the identification of multi-faceted evaluation strategies beyond simple accuracy.
However, formally evaluating a dynamic, interactive clinical situation is hindered by its vast combinatorial space of possible patient states and interaction trajectories, making it difficult to standardize and quantitatively measure such scenarios.
Here, we introduce \textit{AutoMedic}, a multi-agent simulation framework that enables automated evaluation of LLMs as clinical conversational agents.
AutoMedic transforms off-the-shelf static QA datasets into virtual patient profiles, enabling realistic and clinically grounded multi-turn clinical dialogues between LLM agents.
The performance of various clinical conversational agents is then assessed based on our \textit{CARE} metric, which provides a multi-faceted evaluation standard of clinical conversational accuracy, efficiency/strategy, empathy, and robustness.
Our findings, validated by human experts, demonstrate the validity of AutoMedic as an automated evaluation framework for clinical conversational agents, offering practical guidelines for the effective development of LLMs in conversational medical applications.
\end{abstract}

\section{Introduction}
The rapid advancement of large language models (LLMs) has spurred the development of sophisticated foundation models across both proprietary and open-source landscapes~\cite{achiam2023gpt,hurst2024gpt,jaech2024openai,grattafiori2024llama,guo2025deepseek,comanici2025gemini,yang2025qwen3,gpt5}.
This has also led to a significant interest in their medical applications, resulting in a variety of specialized models fine-tuned on clinical data~\cite{chen2023meditron,chen2024huatuogpt,christophe2024med42,OpenBioLLMs,zhang2024ultramedical,jiang2025meds,sellergren2025medgemma}.
Notably, researchers have found from experiments that the knowledge level of these LLMs on standard medical question-answering (QA) benchmarks has become comparable to that of human experts~\cite{singhal2025toward}.

\begin{figure*}[!t]
\centerline{\includegraphics[width=0.99\textwidth]{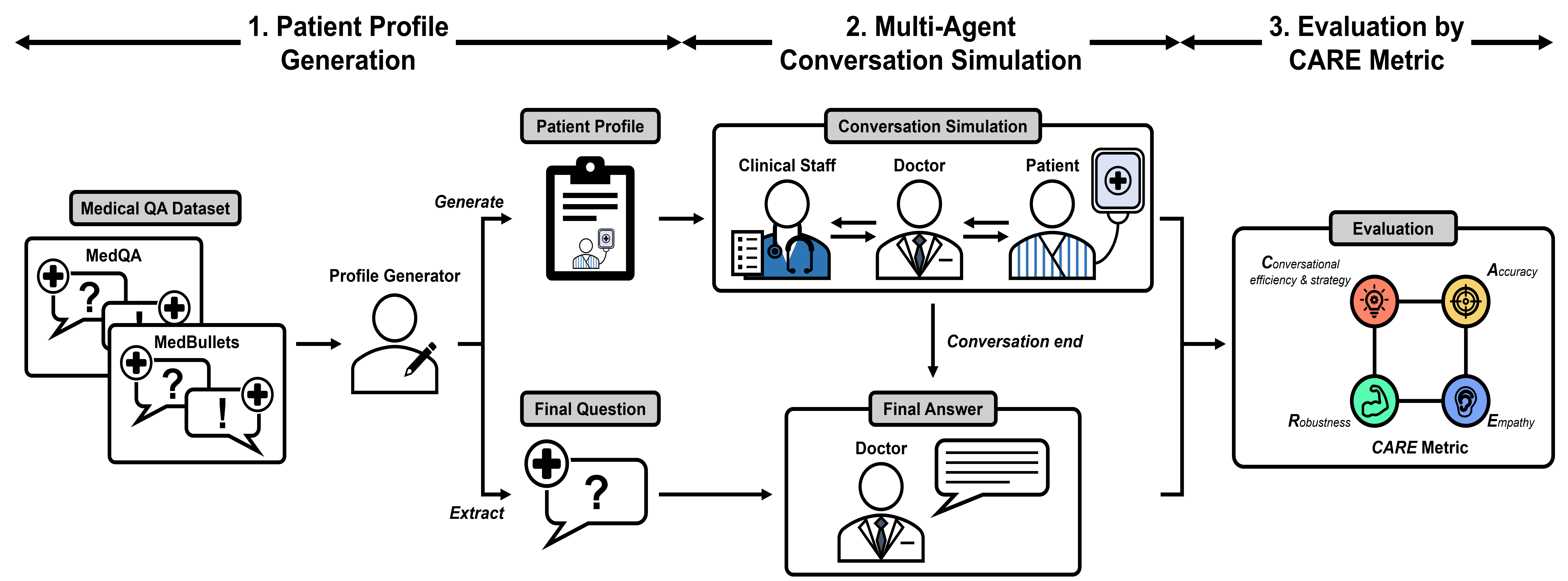}}
\caption{Overview of the proposed AutoMedic framework.
AutoMedic framework automatically converts a medical QA dataset into a virtual patient scenario, facilitates a multi-agent clinical simulation, and evaluates the LLM agent's performance based on our proposed metrics.}
\label{fig:scheme}
\end{figure*}

However, a crucial limitation of evaluations based on existing medical QA benchmarks is their failure to assess the dynamic and interactive nature of real-world clinical encounters.
There are two primary shortcomings.
First, the evaluation process is inherently static.
Unlike a real clinical care scenario where the clinician, who LLMs are expected to take the role in, must interact with the patient to elicit information, these benchmarks provide no measure of conversational and diagnostic inquiry skills.
Effective patient-doctor communication is fundamental to clinical care; it is the primary method for gathering nuanced patient history, building trust, and ensuring patient adherence to treatment plans.
An agent's ability to communicate with empathy and clarity is not just a feature but a core requirement for safe and effective deployment.
Furthermore, given the susceptibility of LLMs to user-driven misinformation or leading inputs~\cite{ziaei2023language,lim2025susceptibility}, the ability to maintain robust and accurate dialogue is critical for safe clinical deployment.
Second, the queries in these benchmarks are typically self-contained, providing all the necessary information upfront.
In actual clinical practice, however, critical information must be actively gathered through patient dialogue and diagnostic tests, which is a vital capability overlooked by current evaluation methods.

To address these limitations, recent studies have made attempts in evaluating LLMs as clinical conversational agents~\cite{schmidgall2024agentclinic,almansooriandkumarMedAgentSim,arora2025healthbench,lee2025psyche,nori2025sequential,tu2025towards}.
These frameworks have demonstrated the feasibility of evaluating LLMs as clinical conversational agents by introducing novel datasets and simulating clinical scenarios.
However, these approaches still present challenges for full automation, as their workflows 1) require specially designed datasets, 2) necessitate substantial human labor or rely on non-realistic clinical simulations, and 3) heavily rely on human evaluators for key qualitative metrics.
Moreover, these methods have notable drawbacks in their extensiveness, often limited to diagnostic scenarios, and lack clear quantitative evaluation metrics.

To address these shortcomings, this paper introduces \textit{AutoMedic}, a fully automated framework for evaluating LLMs as clinical conversational agents in clinical care scenarios, with the overall scheme illustrated in Fig.~\ref{fig:scheme}.
Our key contributions are:
\begin{itemize}
    \item \textbf{An automatic method for converting static medical QA datasets into interactive virtual patient profiles.}
    This approach transforms any standard off-the-shelf QA benchmark into a conversational evaluation suite, removing the need for specialized, hand-crafted datasets.
    Moreover, our framework includes a filtering mechanism that assesses the suitability of source QA items, thereby also enabling an automatic analysis of which datasets are most appropriate for conversion into realistic conversational simulations.
    \item \textbf{A multi-agent simulation framework that models a realistic clinical dialogue.}
    In this framework, the LLM-based clinical conversational agent must actively gather information by conversing with a virtual patient and requesting test results from a clinical staff agent before making a final judgment.
    The entire evaluation process is fully automated, requiring no human intervention.
    \item \textbf{A novel, multi-faceted metric for quantitative evaluation.}
    We propose the \textit{CARE} metric to assess performance across four key rubrics: accuracy, conversational efficiency and strategy, empathy, and robustness.
    By quantifying these crucial aspects, our metric offers a comprehensive guideline for selecting and developing LLMs for real-world clinical deployment.
\end{itemize}

The remainder of this paper is organized as follows.
Section~\ref{sec:related_works} reviews related works on medical LLMs and conversational agent evaluation.
Sections~\ref{sec:framework},~\ref{sec:methods} detail our proposed framework and experimental setup.
Section~\ref{sec:results} and Section~\ref{sec:discussion} present the experimental results and discussion, and Section~\ref{sec:conclusion} concludes the paper.

\section{Related Works}
\label{sec:related_works}
\subsection{LLMs for the Medical Domain}
Leading LLMs have consistently demonstrated high performance on a range of medical tasks and benchmarks.
OpenAI's GPT-4, for example, surpassed the passing score of the United States Medical Licensing Examination (USMLE) by over 20 points~\cite{nori2023capabilities}.
Its successor, GPT-5~\cite{gpt5}, continued this trend by achieving the highest score to date on the HealthBench benchmark~\cite{arora2025healthbench}.
Similarly, DeepSeek-R1~\cite{guo2025deepseek} has shown strong performance in clinical decision-making tasks~\cite{sandmann2025benchmark} and has achieved the top win-rate on several medical benchmarks, outperforming other prominent proprietary and open-source LLMs~\cite{bedi2025medhelm}.

Beyond the aforementioned general-purpose LLMs, a significant amount of research has focused on developing LLMs specifically tailored for the medical domain.
Models such as HuatuoGPT-o1~\cite{chen2024huatuogpt} and m1~\cite{huang2025m1} are fine-tuned on medical problem sets to enhance their clinical reasoning abilities.
In parallel, Google has introduced a series of powerful multimodal medical LLMs, including Med-PaLM M~\cite{tu2024towards}, Med-Gemini~\cite{saab2024capabilities}, and MedGemma~\cite{sellergren2025medgemma}.
These models have demonstrated exceptional performance across a diverse range of medical benchmarks.

However, since most current evaluations primarily assess the knowledge of LLMs via simple question-answering tasks, concerns arise regarding their ecological validity.
Specifically, these methods may fail to adequately gauge a model's utility in real-world clinical settings, which inherently necessitate explorative information gathering for accurate diagnosis and treatment planning~\cite{hager2024evaluation,williams2024evaluating}.

\subsection{Evaluation of Clinical Conversational Agents}
\label{sec:automedic}
\begin{figure*}[!t]
\centerline{\includegraphics[width=0.99\textwidth]{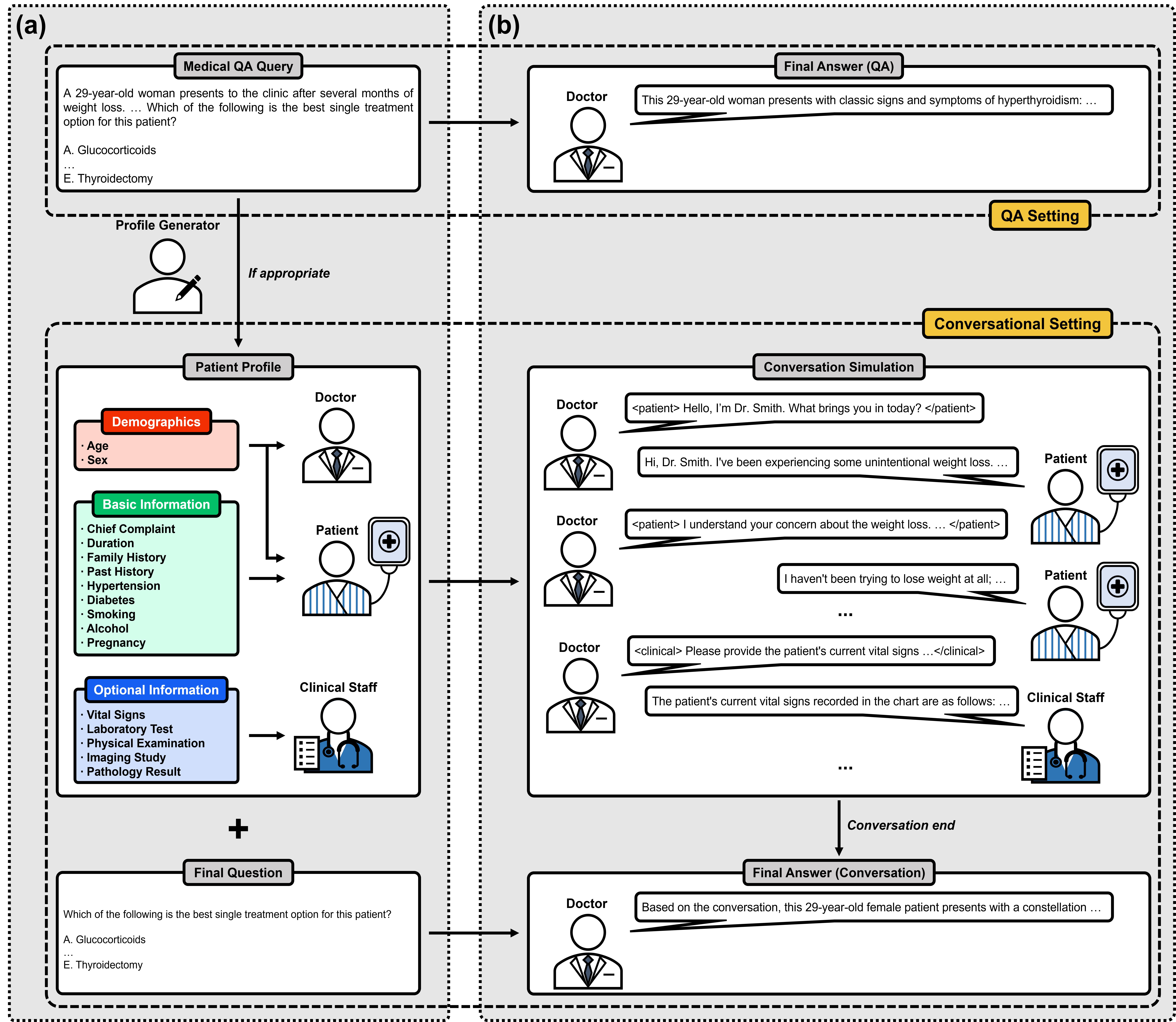}}
\caption{The Automedic framework, detailing (a) the automatic generation of a virtual patient profile from a medical QA query, and (b) the subsequent multi-agent conversation simulation.}
\label{fig:detail}
\end{figure*}

Several pioneering efforts have sought to evaluate LLMs as clinical conversational agents by creating specialized benchmarks and systems.
For instance, Google's Articulate Medical Intelligence Explorer (AMIE), a system for diagnostic dialogue, was evaluated through interactions with human actors playing the role of simulated patients~\cite{tu2025towards}.
Similarly, OpenAI developed HealthBench, a benchmark comprising 5,000 realistic health conversations created in collaboration with physicians, complete with detailed scoring rubrics~\cite{arora2025healthbench}.
More recently, Microsoft introduced the MAI Diagnostic Orchestrator (MAI-DxO), a system where multiple LLM agents collaboratively debate to make an accurate diagnosis.
To assess it, they also released the Sequential Diagnosis Benchmark (SDBench), derived from complex clinicopathological cases from the NEJM~\cite{nori2025sequential}.
A common limitation of these approaches is their dependence on bespoke datasets.
The creation of such resources is a time-consuming and labor-intensive process, which helps explain why the number of conversational benchmarks remains limited compared to the abundance of static medical QA datasets.
Moreover, these datasets are often limited in conversational depth, lacking a significant number of multi-turn exchanges.
Instead of serving as active agents that drive the dialogue, models in these settings are relegated to passive roles, evaluating fixed conversation histories rather than engaging in actual multi-turn interactions.

To overcome these limitations, several studies have specifically focused on evaluating the interactive capabilities of LLM agents in patient dialogue.
A notable example is AgentClinic, a multimodal agent benchmark for evaluating LLMs in simulated clinical environments~\cite{schmidgall2024agentclinic}.
The AgentClinic framework generates virtual patient and measurement agents from medical QA data.
A doctor agent must then interact with these simulated agents to gather the necessary information for a medical decision.
The authors use this simulation to compare the diagnostic accuracy of LLM agents across various biases and languages, supplementing the results with human evaluations of the conversation quality.
Similarly, MedAgentSim~\cite{almansooriandkumarMedAgentSim} simulates doctor-patient interactions but distinguishes itself by utilizing multiple doctor agents to collaboratively derive the final clinical decision.
Despite their comprehensive approach, these approaches has two key limitations relevant to our work.
First, it does not provide a method for assessing whether leveraging an existing medical QA benchmark is suitable for conversion into a realistic conversational simulation.
Second, it lacks a quantitative framework for automatically evaluating an agent's performance across multiple conversational aspects, relying instead on subjective manual human assessment or solely on the accuracy of the QA task.

\section{AutoMedic Framework}
\label{sec:framework}
The AutoMedic framework, illustrated in Fig.~\ref{fig:scheme}, is composed of three primary stages: 1) patient profile generation, 2) multi-agent conversation simulation, and 3) automated evaluation.
This entire process is orchestrated by four specialized LLM agents: a profile generator, a doctor, a patient, and a clinical staff.
The following sections provide the specific roles of these agents and a detailed description of each stage.

\subsection{Agents}
The AutoMedic framework operates using four distinct LLM agents, each with a specialized role.
The primary agent is the \textit{doctor} agent, which emulates a physician and serves as the sole subject of our evaluation.
The other three agents, the \textit{profile generator}, \textit{patient}, and \textit{clinical staff}, act in supporting roles to create the environment for this evaluation.

The process begins with the profile generator, which performs two sequential tasks.
First, it filters a set of medical QA queries to select items suitable for simulation.
Second, it transforms each valid query into a structured virtual patient profile.

This profile is then used by the patient agent, which instantiates the patient's persona.
It dynamically responds to inquiries about symptoms and medical history, and can adapt to a caregiver role in scenarios where a patient cannot self-represent (e.g., due to age or cognitive impairment).

To simulate the broader clinical environment, the clinical staff agent acts as a gatekeeper for technical data.
When the doctor agent requests a test, this agent provides the corresponding results from the profile.
To ensure factual grounding, it only returns pre-existing information and will state that a result is unavailable if it is not in the profile, preventing data hallucination.

Within this environment, the doctor agent's objective is to solve the medical problem by strategically gathering information from both the patient and clinical staff agents.
For baseline comparison, this agent also directly answers the original static QA query without any interaction.

\subsection{Patient Profile Generation}
The first stage of the AutoMedic framework is patient profile generation (Fig.~\ref{fig:detail}(a)).
This process transforms a static medical QA query into a structured virtual patient profile, which then serves as the foundation for the subsequent conversation simulation.
The process begins when the profile generator agent receives a medical QA query, comprising a medical context, a question, several options, and a correct answer, and first assesses its suitability.
As our framework is designed for patient-specific decision-making scenarios, the following types of queries are automatically filtered out and excluded from the dataset by the profile generator agent:
\begin{itemize}
    \item Queries describing research participants or experimental settings rather than clinical care.
    \item Those focusing on abstract concepts (e.g., pathophysiology, molecular mechanisms) instead of a specific patient case.
    \item General knowledge or fact-recall questions (e.g., definitions, classifications).
    \item Image-dependent queries that do not describe the visual findings in the text.
    \item Queries where critical patient details (e.g., age, chief complaint) are only present in the answer choices.
\end{itemize}

Once a query is determined to be suitable, the profile generator agent proceeds to create the virtual patient profile.
It extracts relevant information from the medical context and organizes it into three distinct categories:
\begin{itemize}
    \item \textbf{Demographics}: Basic patient data such as age and sex.
    \item \textbf{Basic Information}: Core clinical details including the chief complaint, duration of symptoms, family and past medical history, and relevant lifestyle factors (e.g., smoking, alcohol use).
    \item \textbf{Optional Information}: Data that may not be present in all cases, such as vital signs, laboratory tests, physical examinations, and imaging results.
\end{itemize}
To enhance the realism of the simulation, the profile generator is instructed to impute any missing basic information items with plausible, randomly generated values, ensuring these additions do not alter the correct answer to the original query.
Conversely, optional information is never generated if it is absent.
This is a crucial constraint, as many queries test the doctor agent's ability to determine which tests are necessary.
Generating these results upfront would interfere with the evaluation.
Finally, the agent isolates the specific medical question and its corresponding response options from the original query, excluding the clinical vignette.
This extracted content is presented to the doctor agent upon the conclusion of the conversation simulation.

Once the patient profile is generated, the information is selectively distributed among the agents according to their role, to simulate realistic knowledge boundaries (Fig.\ref{fig:detail}(b)).
The doctor agent receives only the demographics as the baseline information of the given patient.
The patient agent is provided with both demographics and basic information, enabling it to accurately represent the patient's history and symptoms to answer the query from the doctor agent.
Lastly, the clinical staff agent receives optional information (e.g., examination results) that it provides upon request from the doctor agent.
By isolating technical data until specifically requested, this design maintains the simulation's integrity and ensures the proper evaluation of the doctor agent's information-gathering skills.

\subsection{Multi-Agent Conversation Simulation}
The multi-agent conversation simulation begins after the patient profile has been generated, as illustrated in Fig.~\ref{fig:detail}(b).
The doctor agent initiates and directs the entire dialogue using specific tags.
Questions for the patient agent are wrapped in \texttt{<patient></patient>} tags, while requests for the clinical staff agent are framed within \texttt{<clinical></clinical>} tags.
The patient and clinical staff agents automatically respond to any text directed to them within these respective tags.

This interactive process continues until one of two termination conditions is met: either the doctor agent concludes the dialogue by issuing an \texttt{</end>} tag, or the conversation reaches a predefined maximum length.
We define a single turn as a query from the doctor agent and the corresponding response from another agent.
For our experiments, we set the maximum number of turns to 20, which is sufficient if the doctor agent gathers relevant information efficiently.

Upon the conversation's conclusion, the doctor agent is presented with the final medical question that was extracted from the original query.
The doctor agent then formulates an answer based solely on the information it gathered during the simulated dialogue.
To establish a performance baseline, we also collect responses in a standard QA setting.
As depicted at the top of Fig.~\ref{fig:detail}, the doctor agent is presented with the complete original medical query, containing full patient context upfront.
In this static configuration, the agent must generate an answer without any conversational interaction, mirroring traditional QA tasks.
Both the conversational and static QA responses are subsequently compared during the evaluation phase.

\subsection{Automated Evaluation}
The final stage of the AutoMedic framework is the automated evaluation of the doctor agent.
This evaluation is based on the entire transcript of the multi-agent conversation as well as the final answer provided.
To conduct this assessment automatically and without human intervention, we introduce the \textit{CARE} (Conversation efficiency and strategy, Accuracy, Robustness, and Empathy) metric.
This metric quantitatively assesses the agent's performance across the following four distinct aspects:

\noindent\textbf{1. Accuracy:} The primary goal of this score, $S_{\text{ACC}}$, is to evaluate the diagnostic accuracy of the doctor agent, while also accounting for any performance degradation that occurs when moving from a static to a conversational setting.

First, we independently calculate the accuracy for the conversational setting ($Acc_{\text{Conv}}$) and the static QA setting ($Acc_{\text{QA}}$):
\begin{equation}
    Acc_{\text{Conv}} = \frac{n_{\text{Conv}}}{N}, \quad Acc_{\text{QA}} = \frac{n_{\text{QA}}}{N},
\end{equation}
where $N$ is the total number of queries, and $n_{\text{Conv}}$ and $n_{\text{QA}}$ are the number of correct answers in the conversational and static QA settings, respectively.

The final accuracy score, $S_{\text{ACC}}$, is then defined as:
\begin{equation}\label{eq:acc}
    S_{\text{ACC}} = Acc_{\text{Conv}} \times \frac{Acc_{\text{Conv}}}{ACC_{\text{QA}}}.
\end{equation}
This formula incorporates two components.
The first term, $Acc_{\text{Conv}}$, directly measures the agent's performance in the interactive simulation.
The second term, $Acc_{\text{Conv}}/ACC_{\text{QA}}$, acts as a penalty factor.
It quantifies the drop-off in performance from the ideal-information (static QA) setting to the more challenging conversational setting.
An agent that maintains high accuracy in the conversation relative to its baseline QA performance will score higher.
This ensures that the metric rewards not just correct answers, but also robust conversational information-gathering skills.

\noindent\textbf{2. Conversational Efficiency \& Strategy:} This score, $S_{\text{CES}}$, evaluates the doctor agent's conversational efficiency and its adherence to a natural, strategic dialogue flow.
In real clinical encounters, physicians typically ask one or two focused questions at a time, iteratively building information based on patient responses.
Conversely, some LLM agents may adopt an unnatural ``checklist" approach, asking numerous questions simultaneously to gather information quickly.
While this might appear efficient in terms of raw information acquisition, it is not representative of an effective clinical strategy.

To quantify this, we calculate the average number of words per turn ($w_{\text{turn}}$) used by the doctor agent.
An excessively high $w_{\text{turn}}$ indicates a less natural, multi-question approach within a single turn.
For each sample $i$, the conversational efficiency and strategy score, $s_{\text{CES}}^i$, is assigned as $1 / w_{\text{turn}}$ if the doctor agent correctly answers the final question, and 0 if the answer is incorrect.
This ties conversational style directly to diagnostic success.

The final $S_{\text{CES}}$ score is then computed as the scaled average across all $N$ samples:
\begin{equation}
    S_{\text{CES}} = \frac{1}{N}\sum_{n=1}^{N}s_{\text{CES}}^i \times 100.
\end{equation}
The scaling factor of 100 is applied for clarity and ease of interpretation.

\noindent\textbf{3. Empathy:} The third score, $S_{\text{EMP}}$, assesses the empathetic quality of the doctor agent's communication.
Since LLMs can sometimes interact in a robotic or emotionally detached manner, evaluating their ability to convey empathy is crucial for patient-centric applications.

To quantify this, we use the patient agent as a proxy evaluator.
After each conversation concludes, the patient agent is prompted to rate the doctor agent's empathy on a 5-point scale (where 1 is low and 5 is high).
The final empathy score, $S_{\text{Emp}}$, is the average of these ratings, normalized across all $N$ queries:
\begin{equation}
    S_{\text{EMP}} = \frac{1}{5N}\sum_{n=1}^{N}s_{\text{EMP}}^i,
\end{equation}
where $s_{\text{EMP}}^i$ is the empathy score for the $i$th sample.

\noindent\textbf{4. Robustness}: The final score, $S_{\text{ROB}}$, evaluates the conversational robustness of the doctor agent.
A robust agent must consistently adhere to its designated role and interaction protocols.
Deviations, such as misdirecting questions or breaking character, can compromise the simulation's integrity and lead the conversation astray.

We quantify robustness by identifying and counting specific failure cases during the simulation. A dialogue is considered a failure if any of the following occur:
\begin{itemize}
    \item \textbf{Role-Breaking}: The doctor agent self-simulates the entire interaction, playing the roles of other agents instead of interacting with them.
    \item \textbf{Abrupt Termination}: The conversation ends prematurely (defined as $\leq$ 3 turns) due to the doctor agent's failure to use the required \texttt{<patient>} or \texttt{<clinical>} tags, or due to an improperly placed \texttt {</end>} tag.
    \item \textbf{Invalid Answer}: After a seemingly normal conversation, the doctor agent fails to provide a valid answer to the final question (e.g., by providing an answer not listed in the options).
\end{itemize}

The robustness score, $S_{\text{ROB}}$, is then calculated as the proportion of successful conversations:
\begin{equation}
    S_{\text{ROB}} = 1 - \frac{n_{\text{Fail}}}{N}
\end{equation}
where $n_{\text{Fail}}$ is the total count of the failure cases described above.

\section{Methods}
\label{sec:methods}
\subsection{Datasets}
To evaluate the performance of each LLM as a clinical conversational agent using the AutoMedic framework, we utilized six distinct medical QA datasets.

\textbf{MedBullets}: This dataset~\cite{chen2025benchmarking} comprises 308 multiple-choice questions formatted in the style of the USMLE, each with five answer options.

\textbf{MedQA}: Another USMLE-style medical QA benchmark, MedQA~\cite{jin2021disease} contains 1,273 multiple-choice questions, each presenting five answer options.

\textbf{MedXpertQA}: MedXpertQA~\cite{zuo2025medxpertqa} is a medical QA benchmark featuring both textual and multimodal questions.
Our evaluation focused exclusively on its text-based subset, which consists of 2,450 questions, each with ten answer choices.

\textbf{MedMCQA}: MedMCQA~\cite{pal2022medmcqa} is a multiple-choice question answering dataset derived from real-world medical entrance exams.
For our experiments, we used its validation split, which provides the correct answers.
After filtering to include only single-choice questions, our subset contained 2,816 questions, each with four answer options.

\textbf{HEAD-QA}: This multi-choice healthcare dataset~\cite{vilares-gomez-rodriguez-2019-head} includes medical questions in both Spanish and English.
We specifically used the test split of its English subset, which provides 2,742 questions, each with four answer options.

\textbf{MMLU-Pro}: MMLU-Pro~\cite{wang2024mmlu} is a multi-task understanding dataset designed for rigorous LLM benchmarking.
From this dataset, we extracted questions pertaining only to the health and biology categories, totalling 1,535 questions with three to ten answer options each.

\begin{table}[!t]
    \centering
    \caption{List of LLMs included in our experiments}
    \label{tab:model}
    \resizebox{0.9\columnwidth}{!}{
    \begin{tabular}{c|c|c|c}
    \hline
    Name    & Type  & Domain    & Size  \\
    \hline\hline
    Llama 3 & \multirow{7}{*}{Open-Source}  & \multirow{4}{*}{General}  & 70B   \\
    \cline{1-1}\cline{4-4}
    Qwen3   &   &   & 32B   \\
    \cline{1-1}\cline{4-4}
    DeepSeek-R1 &   &   & 70B   \\
    \cline{1-1}\cline{4-4}
    gpt-oss &   &   & 120B  \\
    \cline{1-1}\cline{3-4}
    Med42-v2    &   & \multirow{3}{*}{Biomedical}   & 70B   \\
    \cline{1-1}\cline{4-4}
    OpenBioLLM  &   &   & 70B \\
    \cline{1-1}\cline{4-4}
    HuatuoGPT-o1    &   &   & 72B \\
    \hline
    GPT-4o  & \multirow{2}{*}{Proprietary}  & \multirow{2}{*}{General}  & - \\
    \cline{1-1}\cline{4-4}
    Claude Sonnet 4 &   &   & - \\
    \hline
    \end{tabular}
    }
\end{table}

\subsection{Models}
For the doctor agent, we evaluate the performance of 11 distinct LLMs, which are detailed in TABLE~\ref{tab:model}.
These models are grouped into three categories.

\textbf{Open-Source General LLMs}: We selected a range of general-purpose open-source models.
We utilized the instruction-tuned versions of Llama 3~\cite{grattafiori2024llama} and Qwen3~\cite{yang2025qwen3}.
As a representative reasoning model, we included the 70B distilled version of DeepSeek-R1~\cite{guo2025deepseek}.
We also used gpt-oss~\cite{agarwal2025gpt}, another open-source reasoning model from OpenAI, in its 120B parameter version.

\textbf{Open Source Biomedical LLMs}: To assess domain-specific models, we employed two LLMs designed for the biomedical field: Med42-v2~\cite{christophe2024med42} and OpenBioLLM~\cite{OpenBioLLMs}, both with 70B parameters.
We also included HuatuoGPT-o1~\cite{chen2024huatuogpt}, a 72B parameter model specifically tailored for medical reasoning tasks.

\textbf{Proprietary General LLMs}: Finally, we evaluated two high-performing proprietary models: GPT-4o~\cite{hurst2024gpt} and Claude Sonnet 4~\cite{claude4}.

For the other agents (profile generator, patient, and clinical staff), we employed GPT-4o.
Although other state-of-the-art LLMs could be used for these roles and their impact could be analyzed, the primary aim of this study is to evaluate the performance of the doctor agent.
Therefore, GPT-4o was used for all supporting agents to ensure a stable and consistent conversation simulation.

\subsection{Human Evaluation}
To validate the accuracy and effectiveness of our framework, we conducted three distinct human expert evaluations with the participation of four licensed medical professionals.
These evaluations validate the clinical validity of the AutoMedic framework and the CARE metric by demonstrating their alignment with the judgments of medical experts.
The institutional review board approved these human evaluation studies (IRB No. 4-2025-0982).

The first study assessed the accuracy of the profile generator's filtering mechanism.
We randomly sampled 24 medical queries from the six datasets, comprising 12 queries classified as "appropriate" and 12 as "inappropriate" by the profile generator agent.
The human experts then independently judged the suitability of each sample for patient profile generation.
We measured inter-rater reliability among the experts using Fleiss' Kappa and the agreement between our agent's classification and the experts' majority vote using Cohen's Kappa.

The second study evaluated the quality of the generated patient profiles.
Experts were asked to rate 21 randomly sampled profiles on a 4-point scale.
The evaluation criteria included faithfulness to the original medical query, completeness of information, and the clinical plausibility of any imputed data.
To measure inter-rater reliability, we calculated both the percent agreement and Gwet's AC2 coefficient, a metric robust to chance agreement.
We then determined the average quality score across all profiles.

The final study aimed to correlate our automated CARE metric with human judgment.
We generated 30 simulation results (15 each for Llama 3-70B and Claude Sonnet 4 as the doctor agent on the MedQA dataset).
In a blinded setup, the human experts rated each result on a 3-point scale across the four CARE dimensions: accuracy, conversational efficiency and strategy, empathy, and robustness.
We measured inter-rater reliability using both percent agreement and Gwet's AC2 coefficient.
Finally, we compared the trend of the experts' average scores against the trend of the calculated CARE metric scores for each model.

\section{Results}
\label{sec:results}
\subsection{Human Evaluation of Patient Profile Generation.}
\begin{table}[!t]
    \centering
    \caption{Human Expert Evaluation Results for Patient Profile Generation.
    }
    \label{tab:profile}
    \resizebox{0.9\columnwidth}{!}{
    \begin{tabular}{c|c}
    \hline
    \multicolumn{2}{c}{Filtering Accuracy of the Profile Generator} \\
    \hline
    Fleiss' Kappa    & 0.7630    \\
    Cohen's Kappa    & 0.8197    \\
    \hline\hline
    \multicolumn{2}{c}{Quality of Generated Patient Profiles} \\
    \hline
    Percent Agreement   & 0.9140    \\
    Gwet's AC2          & 0.8313    \\
    \hline
    Average Score       & 3.464    \\
    \hline
    \end{tabular}
    }
    \begin{minipage}{0.95\columnwidth}
    \vspace{0.1cm}
    \small Notes: For the filtering accuracy assessment, Fleiss' Kappa quantifies inter-rater agreement among human experts, while Cohen's Kappa measures agreement between the experts' majority vote and the profile generator.
    For the quality evaluation of generated patient profiles, the percent agreement and Gwet's AC2 indicate human expert reliability, and the ``Average Score'' represents the mean expert rating across all samples (maximum possible score: 4 points).
    \end{minipage}
\end{table}

To justify the patient profile generation process, we conducted two human expert evaluations.
The first specifically assessed the profile generator's filtering accuracy.
Human experts demonstrated substantial agreement on the suitability of queries for profile generation, indicated by a Fleiss' Kappa of 0.7630 (TABLE~\ref{tab:profile}).
This high inter-rater reliability confirms that our filtering criteria are well-defined and consistently applied.
Furthermore, the agreement between the profile generator's classifications and the human experts' majority vote was almost perfect, yielding a Cohen's Kappa of 0.8197 (excluding two tied cases).
This evaluation robustly demonstrates that our profile generator accurately identifies and filters out inappropriate medical queries before patient profile generation.

Next, the evaluation of the generated patient profiles further supported our framework's design.
Human expert ratings showed strong reliability, evidenced by a percent agreement of 0.9140 and a Gwet's AC2 coefficient of 0.8313 (TABLE~\ref{tab:profile}).
This indicates consistent expert judgment.
Additionally, the profiles received an average score of 3.464 out of a possible 4 points.
This high average score suggests that the patient profiles generated by our framework are generally accurate and clinically reliable.
These evaluations collectively validate the effectiveness of our patient profile generation process.

\subsection{Dataset Suitability for Clinical Simulation}
\begin{figure}[!t]
\centerline{\includegraphics[width=0.95\columnwidth]{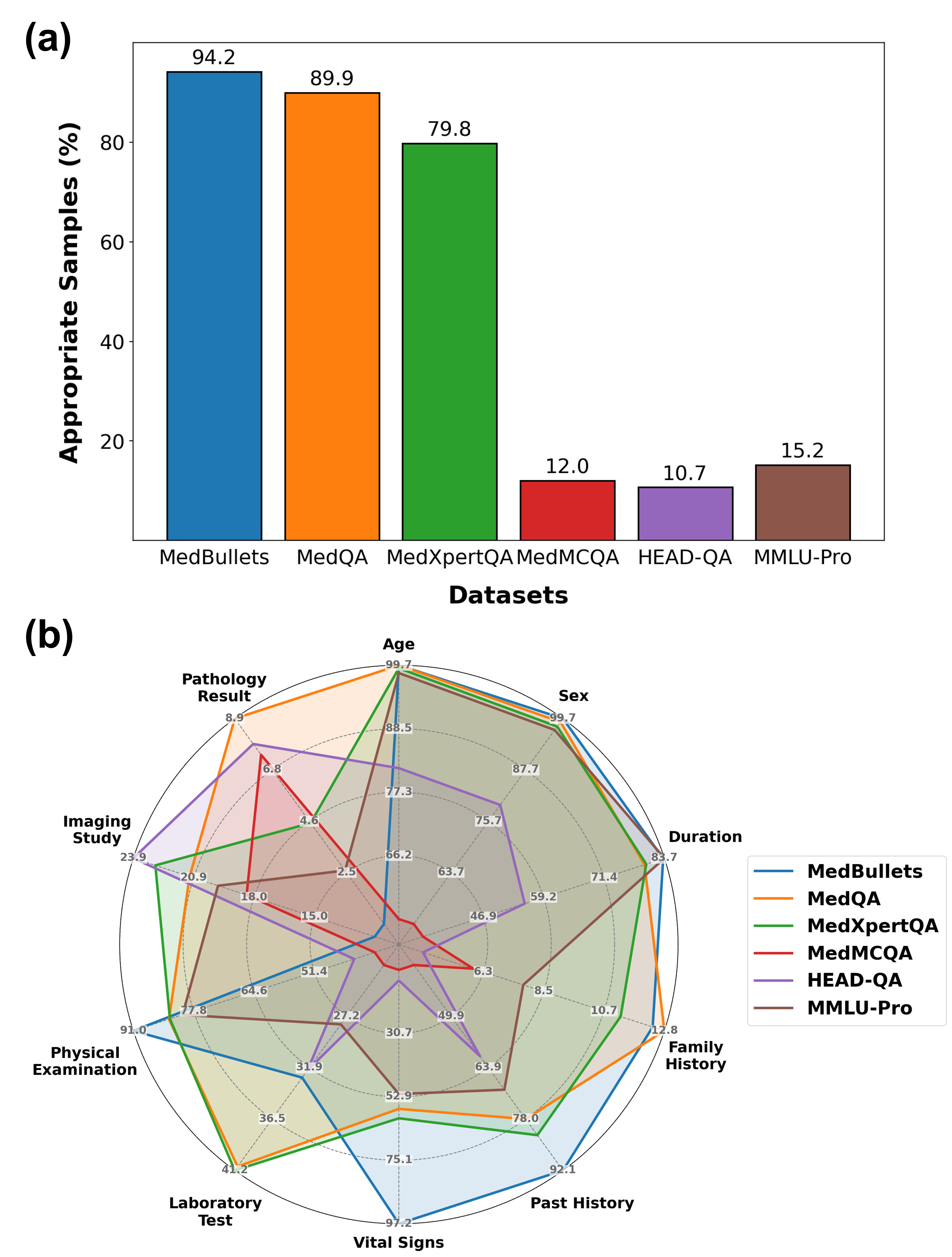}}
\caption{Analysis of Medical QA Benchmark Datasets for Clinical Simulation.
(a) A bar chart illustrates the proportion of ``Appropriate Samples'' within each dataset (calculated from the total samples).
(b) A radar chart illustrates the percentage of appropriate samples that contain specific demographic, basic, and optional information for patient profile generation (calculated from the appropriate samples).}
\label{fig:dataset}
\end{figure}

Before evaluating the performance of LLMs with our AutoMedic framework, we conducted a pre-analysis of the selected medical QA benchmarks to assess their suitability for simulating realistic clinical situations.
Leveraging our validated patient profile generation process, we analyzed each dataset based on its capacity to provide sufficient and relevant information for virtual patient profiles.
The detailed findings are presented in Fig.~\ref{fig:dataset}.

Our analysis first focused on the proportion of samples deemed appropriate for generating a virtual patient profile.
As illustrated in Fig.~\ref{fig:dataset}(a), MedBullets and MedQA exhibited the highest proportion, with over 90\% of their samples being suitable.
MedXeprtQA, while having a slightly lower ratio of appropriate samples (79.8\%), still contributed the largest absolute number of appropriate samples (1,955).
In contrast, MedMCQA, HEAD-QA, and MMLU-Pro showed a significantly lower proportion of appropriate samples, ranging from 10\% to 15\%.
This initial filter is crucial, as only appropriate queries can lead to meaningful clinical simulations.

\begin{table*}[!t]
    \centering
    \caption{Alignment of Human Expert and CARE Metric Evaluation for LLMs as Doctor Agents.
    }
    \label{tab:align}
    \resizebox{0.9\textwidth}{!}{
    \begin{tabular}{c|c|c|c|c|c}
    \hline
    \multicolumn{2}{c|}{}   & \multirowcell{2}{Accuracy} & \multirowcell{2}{Conversational\\Efficiency \& Strategy}    & \multirowcell{2}{Empathy}    & \multirowcell{2}{Robustness}  \\
    \multicolumn{2}{c|}{} & &   &   &   \\
    \hline\hline
    \multicolumn{2}{c|}{Percent Agreement}   & 0.9722    & 0.8185    & 0.9185    & 0.7426    \\
    \multicolumn{2}{c|}{Gwet's AC2}  & 0.9460    & 0.6002    & 0.8571    & 0.6209    \\
    \hline
    \multirow{2}{*}{Average Human Score}    & Claude Sonnet 4   & 2.2333   & 2.5000   & 2.9333   & 2.6500   \\
    \   & Llama 3-70B   & 1.6333    & 2.0167    & 2.3667    & 2.7333    \\
    \hline
    \multirow{2}{*}{CARE Metric}    & Claude Sonnet 4   & 0.4699    & 1.0192    & 0.8520    & 0.9333    \\
    \   & Llama 3-70B   & 0.2866    & 0.6703    & 0.7260    & 1.0000    \\
    \hline
    \end{tabular}
    }
    \begin{minipage}{0.95\textwidth}
    \vspace{0.1cm}
    \small Notes: This table presents inter-rater reliability metrics (Percent Agreement and Gwet's AC2), average human expert scores (3-point Likert scale), and corresponding CARE metric scores for Claude Sonnet 4 and Llama 3-70B across four evaluation dimensions when acting as a doctor agent on the MedBullets dataset.
    \end{minipage}
    \vspace{-0.5cm}
\end{table*}

Next, we assessed the informational richness of the appropriate samples, specifically evaluating whether the datasets provided sufficient details to construct comprehensive patient profiles.
As shown in Fig.~\ref{fig:dataset}(b), demographics, such as age and sex, were generally well-covered across most datasets, though MedMCQA provided this information for only about half of its appropriate samples.
For basic clinical information like symptom duration and past medical history, MedBullets, MedQA, and MedXpertQA consistently included these details for over 70\% of their samples.
Conversely, MedMCQA and HEAD-QA showed noticeable gaps, providing this information for a relatively smaller fraction of their queries.
This disparity became even more pronounced when examining optional information, such as vital signs or physical examination findings.
MedMCQA and HEAD-QA again presented the lowest ratios of inclusion.
Interestingly, MMLU-Pro, despite its low proportion of appropriate samples, provided patient information with a frequency similar to the more suitable datasets once a sample was deemed appropriate.

In summary, MedBullets, MedQA, and MedXpertQA emerge as the most suitable datasets for the Automedic framework.
They not only offer a high proportion (or a large absolute number) of appropriate samples for patient profile generation but also provide abundant and comprehensive information to create detailed virtual patient profiles, essential for accurate clinical situation simulation.
MedBullets and MedQA stand out for their high proportional suitability, while MedXpertQA excels in the total volume of usable samples.

Conversely, MedMCQA and HEAD-QA, despite potentially having large total numbers of medical queries, are less optimal for our framework due to their low proportion of appropriate samples and their subsequent lack of detailed information necessary for robust patient profile generation.
MMLU-Pro, while also having a low proportion of appropriate samples, is somewhat mitigated by the relatively good information completeness when a sample is appropriate, suggesting its suitable samples are of higher quality.

\subsection{Correlation of CARE Metric with Human Expert Evaluations}
Before deploying the CARE metric for LLM evaluation, we first established its alignment with the human expert judgment.
TABLE~\ref{tab:align} presents the results of human expert evaluation for Cluade Sonnet 4 and Llama 3-70B acting as doctor agents on a random sample from the MedQA dataset, alongside their corresponding CARE metric scores.

For accuracy, human experts demonstrated very high reliability, with a Gwet's AC2 coefficient of 0.9460.
This high agreement likely stems from the relatively objective nature of judging the correctness of a medical answer.
Claude Sonnet 4 received a notably higher average score of 2.2333 compared to Llama 3-70B's 1.6333.
This trend is directly reflected in the CARE metric, where Claude Sonnet 4's $S_{\text{ACC}}$ (0.4699) is also substantially higher than Llama 3-70B's (0.2866), indicating strong alignment.

Regarding conversation efficiency and strategy, the human expert agreement, while lower than for accuracy, remained substantial with a Gwet's AC2 of 0.6002.
Human evaluators rated Claude Sonnet 4 at 2.5000, slightly higher than Llama 3-70B's 2.0167.
This performance hierarchy is consistently mirrored by the $S_{\text{CES}}$ score of the CARE metric, with Claude Sonnet 4 scoring 1.0192 and Llama 3-70B scoring 0.6703.
This suggests that the automated metric effectively captures the nuances of conversational flow and strategy perceived by human experts.

For empathy, human evaluators exhibited high agreement (0.9185 percent agreement and 0.8571 Gwet's AC2 coefficient).
Claude Sonnet 4 achieved a near-perfect average human score of 2.9333, significantly outperforming Llama 3-70B, which scored 2.3667.
The $S_{\text{EMP}}$ score of the CARE metric also reflects this hierarchy, with Claude Sonnet 4 (0.8520) demonstrating higher empathy scores than Llama 3-70B (0.7260), further reinforcing the alignment.

Finally, for robustness, human experts showed substantial agreement, with a Gwet's AC2 coefficient of 0.6209.
Interestingly, both models received similarly high average human scores, with Llama 3-70B (2.7333) slightly edging out Claude Sonnet 4 (2.6500).
This trend is compellingly aligned with $S_{\text{ROB}}$ score, where Llama 3-70B achieved a perfect score of 1.0000, and Claude Sonnet 4 also scored highly at 0.9333.
This high performance for both models suggests that even under potentially challenging conversational conditions, both LLMs maintained a high degree of consistency in their responses.

In summary, the high inter-rater reliability observed across all four CARE categories underscores the robustness and consistency of the human evaluation results.
Crucially, the trends in average human expert scores show strong alignment with the corresponding CARE metric values regardless of LLMs.
This robust correlation validated the CARE metric as an appropriate and reliable automated tool for evaluating the multifaceted capabilities of LLMs as clinical conversational agents.

\subsection{Accuracy in Static and Conversational Settings}
\begin{figure*}[!t]
\centerline{\includegraphics[width=0.99\textwidth]{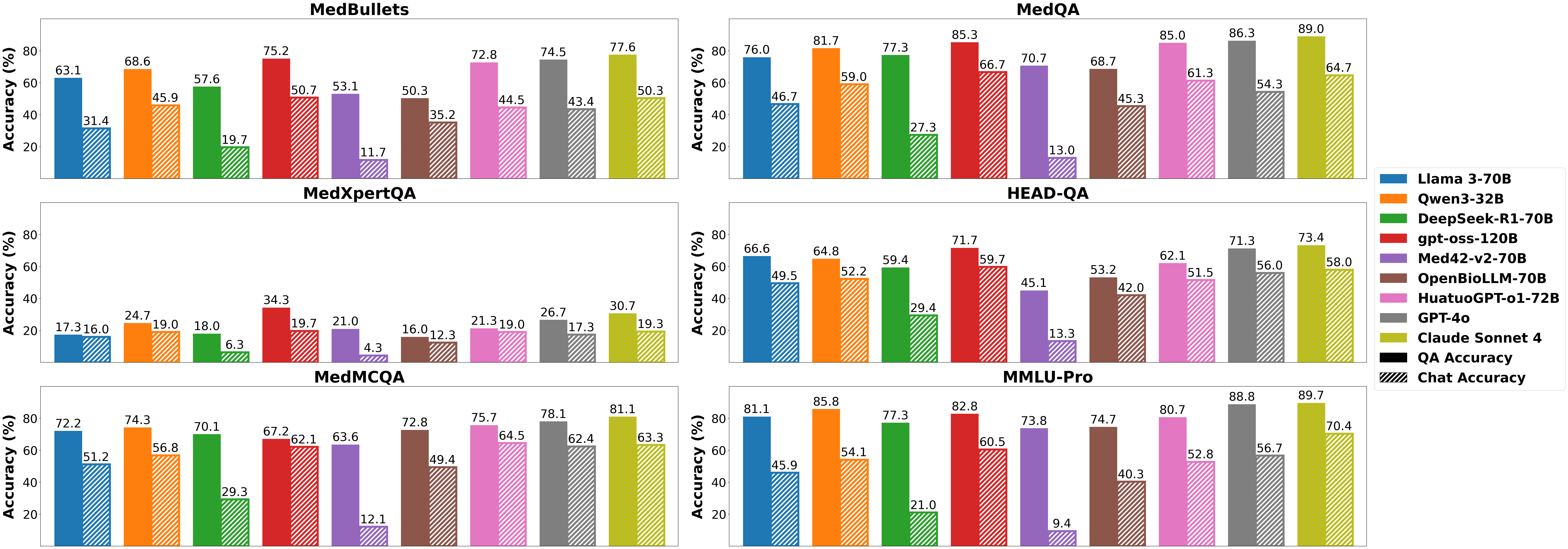}}
\caption{Static and conversational QA accuracy of LLMs across medical QA benchmarks.
This figure presents the static QA accuracy (solid bars) and conversational QA accuracy (striped bars) for each LLM across the six medical QA benchmark datasets.}
\label{fig:accuracy}
\end{figure*}

First, we comprehensively compare the performance of LLMs by examining their accuracy in both static QA and conversational settings.
Fig.~\ref{fig:accuracy} illustrates these accuracies across the various medical QA datasets.

A clear trend emerges whereby models exhibiting high QA accuracy generally maintain relatively high conversational accuracy.
However, a consistent observation across all experiments is that every model demonstrates lower performance in the conversational setting compared to its static QA counterpart.
This performance drop is particularly pronounced for DeepSeek-R1-70B and Med42-v2-70B, as evident in their significant accuracy decline in conversational scenarios across most datasets.
This universal struggle suggests that models, regardless of their underlying knowledge, face substantial challenges in effectively gathering all necessary information within a dynamic conversational exchange.

This finding provides a critical insight that
an LLM's raw knowledge in medicine (reflected in QA accuracy) does not directly translate to effective clinical conversational ability.
The conversational context introduces complexities such as managing dialogue flow, interpreting implicit cues, maintaining coherence, and adapting to user responses, all of which are distinct from simply answering a well-posed, single-turn question.
Therefore, evaluating LLMs specifically as clinical conversational agents, rather than solely on static QA, is essential.
This approach reveals whether models can effectively communicate with users, even if they possess extensive domain knowledge, highlighting the importance of metrics that capture interactive performance.

\subsection{Clinical Conversation Performance Analysis using the CARE Metric}
\begin{figure*}[!t]
\centerline{\includegraphics[width=0.99\textwidth]{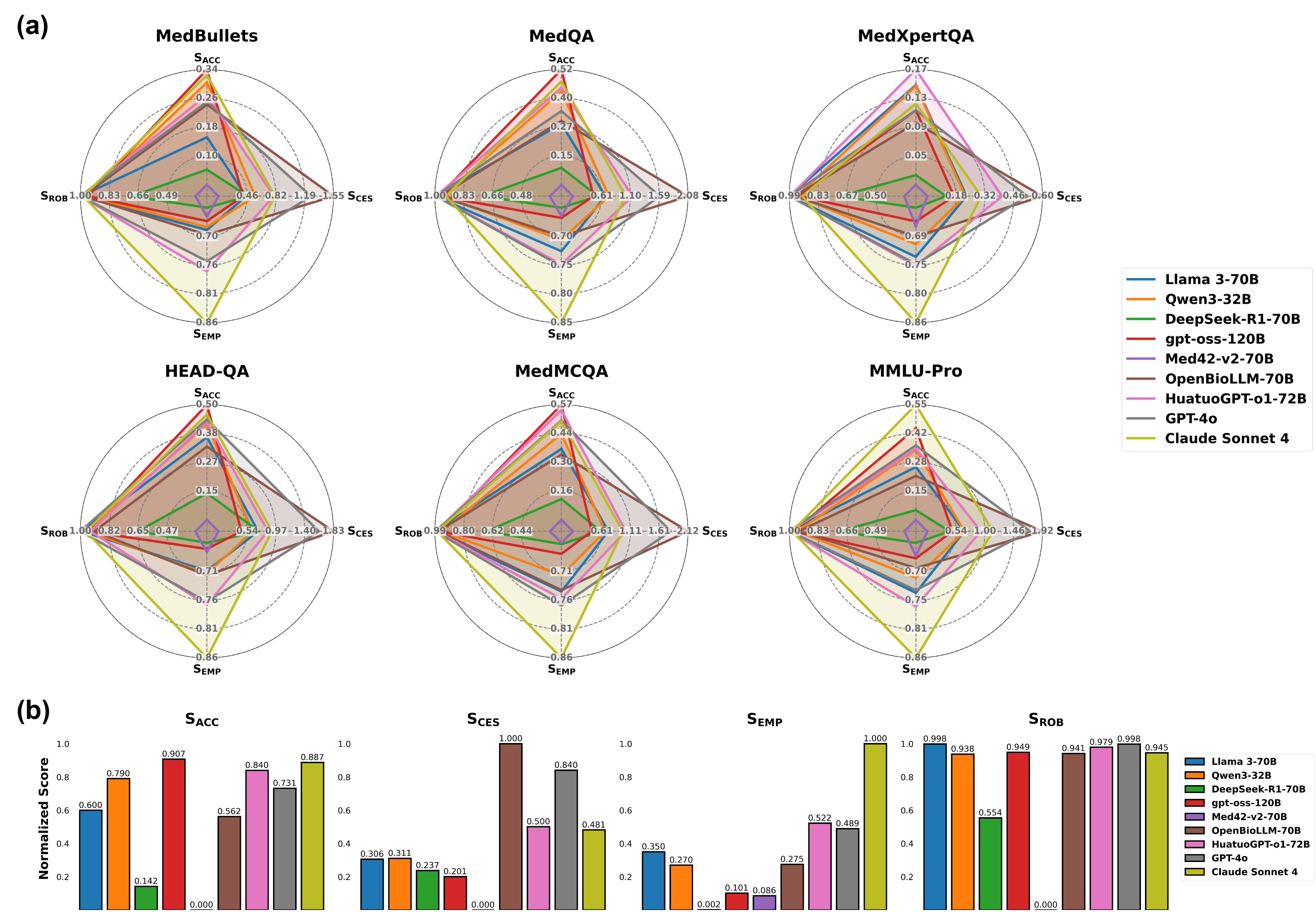}}
\caption{(a) LLM performance across medical QA benchmarks via CARE metric.
These radar charts display the performance of various LLMs on each medical QA dataset, illustrating scores across the four CARE metric dimensions: $S_{\text{ACC}}$ (accuracy), $S_{\text{CES}}$ (conversational efficiency \& strategy), $S_{\text{EMP}}$ (empathy), and $S_{\text{ROB}}$ (robustness).
(b) Average normalized CARE metric across medical QA benchmarks.
This bar plot illustrates the average normalized scores for each component of the CARE metric ($S_{\text{ACC}}$, $S_{\text{CES}}$, $S_{\text{EMP}}$, $S_{\text{ROB}}$) across all evaluated medical QA datasets.
Scores are min-max normalized per dataset, with 1 representing the highest score and 0 the lowest for each metric.}
\label{fig:metric}
\end{figure*}

Next, we evaluated the capabilities of various LLMs as clinical conversational agents using the comprehensive CARE metric.
Fig.~\ref{fig:metric}(a) visually summarizes these evaluation results across the medical QA datasets.

Among the open-source general models, Llama 3-70B and Qwen3-32B exhibited average performance across the overall CARE metric.
Llama3-70B notably achieved the highest $S_{\text{ROB}}$ across the board, though the inter-model differences in robustness were generally minor.
Qwen3-32B demonstrated strong $S_{\text{ACC}}$, yet it did not surpass larger models like gpt-oss-120B or the proprietary models.
DeepSeek-R1-70B consistently performed poorly across most experiments.
As illustrated in Figure~\ref{fig:example_deepseek_gptoss} in the Appendix, DeepSeek-R1-70B struggled to function appropriately within a virtual clinical setting (e.g., ended the conversation inappropriately).
We hypothesize this degradation in performance stems from its distilled nature, where supervised fine-tuning on synthetic data, rather than reinforcement learning, may have impaired both its reasoning capabilities and its ability to engage in proper communication with other agents, a phenomenon observed in prior work~\cite{oh2025rethinking}.
gpt-oss-120B achieved high $S_{\text{ACC}}$ due to strong QA and conversational accuracy, but received low scores for both $S_{\text{CES}}$ and $S_{\text{EMP}}$.
This deficiency in conversational strategy and empathy is due to the model's reliance on uncharacteristic formats, such as numbered lists (Figure~\ref{fig:example_deepseek_gptoss} in the Appendix), a style that emphasizes objective communication over human interaction. This detached, objective approach inherently ensures the absence of sycophancy.

Turning to open-source biomedical LLMs, Med42-v2-70B generally performed poorly across most CARE metrics.
Despite being fine-tuned on medical data, it showed no significant improvement in $S_{\text{ACC}}$ and underperformed in other dimensions.
Moreover, as shown in Figure~\ref{fig:example_med42_claude} in the Appendix, it exhibited abnormal conversational behavior, such as answering its own questions without waiting for input form the patient or clinical staff.
OpenBioLLM-70B achieved the best $S_{\text{CES}}$ and better $S_{\text{ROB}}$ compared to Med42-v2-70B, but it too exhibited degraded performance in accuracy and empathy.
In contrast, HuatuoGPT-o1-72B demonstrated superior performance across all CARE metrics.
Its $S_{\text{ACC}}$ scores were comparable to those of gpt-oss-120B and proprietary models, which we attribute to its combination of medical domain-specific tuning and robust reasoning abilities.

\begin{figure*}[!t]
\centerline{\includegraphics[width=0.99\textwidth]{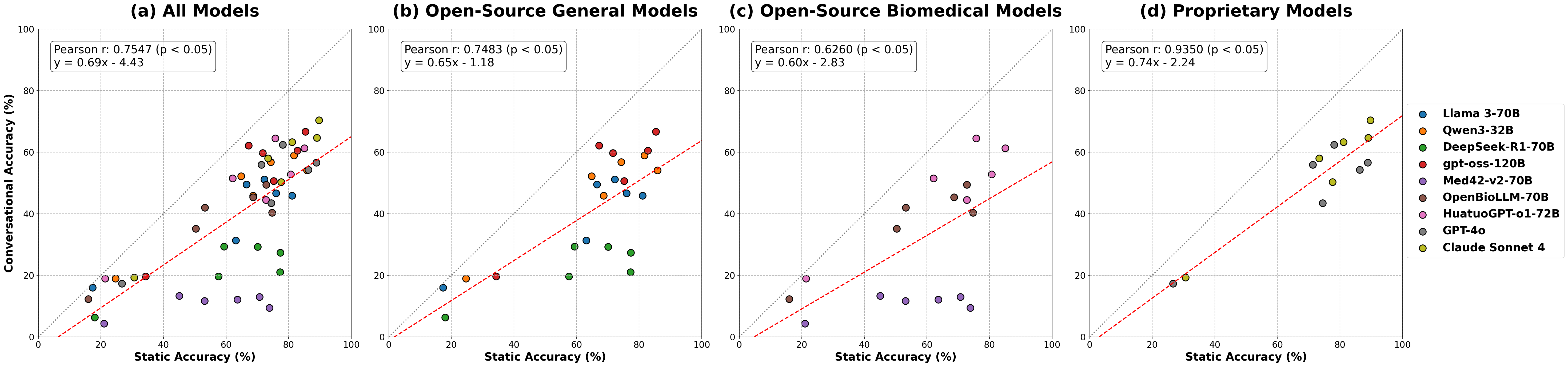}}
\caption{The relationship between static and conversational QA accuracy.
The scatter plots display the correlation for: (a) All models, (b) Open-source general models, (c) Open-source biomedical models, and (d) Proprietary models.
For each category, the Pearson correlation coefficient ($r$), $p$-value, and linear regression equation are provided in the top-left corner.
The grey dashed line represents the identity line ($y=x$), while the red dashed line indicates the fitted regression line.}
\label{fig:qa_chat_scatter}
\end{figure*}

The two proprietary general LLMs presented different strengths.
Claude Sonnet 4 achieved a high $S_{\text{ACC}}$ due to its excellent QA and conversational accuracy, with a minimal gap between the two.
Furthermore, it recorded the highest $S_{\text{EMP}}$ across all models, consistently exceeding 0.85 on all datasets.
This superior empathetic performance is likely due to its tendency to check patient status and express empathy, as shown in Figure~\ref{fig:example_med42_claude} in the Appendix, unlike other LLMs that primarily focus on information gathering.
GPT-4o, while showing relatively lower $S_{\text{ACC}}$ and $S_{\text{EMP}}$ compared to Claude Sonnet 4, slightly outperformed it in $S_{\text{ROB}}$.
This indicates that GPT-4o possesses superior robustness, effectively maintaining its assigned persona and adhering to role constraints throughout extended multi-turn conversations.
Moreover, GPT-4o exhibited superior $S_{\text{CES}}$, indicating its efficiency in gathering necessary patient information with minimal turns and words.

To enable a comprehensive comparison of model performance across diverse datasets, we applied min-max normalization for each metric within each dataset (where 1 represents the highest score and 0 the lowest).
Subsequently, these normalized scores were averaged across all datasets for each metric.
Fig.~\ref{fig:metric}(b) presents these average normalized overall CARE metric scores.

For $S_{\text{ACC}}$, gpt-oss-120B demonstrates the strongest performance, closely followed by Claude Sonnet 4 and HuatuoGPT-o1-72B.
In terms of $S_{\text{CES}}$, OpenBioLLM achieves the highest average score, with GPT-4o securing the second position.
Regarding $S_{\text{EMP}}$, Claude Sonnet 4 stands out with an overwhelmingly superior performance, indicating its exceptional ability in this dimension.
For $S_{\text{ROB}}$, Llama 3-70B leads, although the performance differences among models in this category are relatively small.
Notably, DeepSeek-R1-70B and Med42-v2-70B consistently exhibit poor performance across most CARE metrics.
These specific trends will be discussed in greater detail in the subsequent discussion section.

\section{Discussion}
\label{sec:discussion}
In this study, we introduced AutoMedic, a novel agentic framework designed for the fully automated evaluation of LLMs as clinical conversational agents.
To enable accurate and automated evaluation of LLMs, we proposed the CARE metric, which enables a multi-faceted assessment of LLMs across four critical dimensions: accuracy, conversational efficiency and strategy, empathy, and robustness.
Through extensive human expert evaluations, we validated the reliability of the virtual patient profile generation process within AutoMedic and demonstrated that our CARE metric closely aligns with the clinical judgment of human experts.

Leveraging the patient profile generation capabilities of AutoMedic, we first established guidelines for assessing the suitability of medical QA datasets for clinical simulation.
Our analysis revealed significant disparities among datasets regarding the volume of clinically appropriate samples and the depth of patient information provided.
We found that datasets such as MedBullets, MedQA, and MedXpertQA offer a substantial number of appropriate samples enriched with sufficient patient details.
In contrast, benchmarks like MedMCQA and HEAD-QA exhibit a low proportion of suitable samples and frequently lack basic patient information.
These results imply that not all medical QA datasets contain questions relevant to realistic clinical situations.
Therefore, relying solely on them to assess the clinical capabilities of LLMs may be inadequate.
Consequently, our framework serves as a vital guideline for determining the applicability of medical QA datasets to clinical scenario simulations.
Furthermore, a key advantage of our framework is its generalizability.
Although we utilized six specific datasets in this study, AutoMedic is designed to automatically adapt to and process any new off-the-shelf medical QA dataset without manual modification.

We compared the performance in both static QA and conversational settings, revealing that the dynamic conversational environment presents a significantly greater challenge than static QA.
To quantify the relationship between these two performance modes, we performed a linear regression analysis and calculated the Pearson correlation coefficient ($r$).
As shown in Fig.~\ref{fig:qa_chat_scatter}(a), a strong positive correlation exists between static and conversational accuracy across all models ($r$=0.7547).
When examining open-source general models (Fig.~\ref{fig:qa_chat_scatter}(b)), a similar strong correlation is observed, although DeepSeek-R1-70B notably deviates from the general trend.
In contrast, open-source biomedical models (Fig.~\ref{fig:qa_chat_scatter}(c)) exhibit the weakest correlation ($r=0.6260$).
These results suggest that while static medical knowledge serves as a foundation for conversational competence, suboptimal fine-tuning strategies may hinder the effective translation of this knowledge into clinical dialogue.
Conversely, proprietary models (Fig.~\ref{fig:qa_chat_scatter}(d)) demonstrate the highest correlation ($r=0.9350$), indicating a consistent balance between their static clinical knowledge and conversational capabilities.
In conclusion, for clinical conversational settings, it is advantageous to prioritize proprietary models or carefully select open-source models that demonstrate both high static QA accuracy and robust instruction-following tuning.

Our evaluation using the CARE metric revealed distinct performance profiles for each LLM, highlighting unique strengths and weaknesses.
One particularly striking finding was that domain-specific medical tuning did not confer superiority in conversational settings.
With the notable exception of HuatuoGPT-o1-72B, most medical-tuned models performed worse than their general-domain counterparts.
We conjecture that this performance deficit arises because these models were primarily fine-tuned for static medical QA tasks, not for dynamic, interactive communication.
During such specialized fine-tuning, the models may inadvertently lose their broader communicative capabilities.
Consequently, when deployed in realistic clinical scenarios, they fail to leverage their specialized knowledge effectively because they cannot manage the conversational interaction.
This result strongly indicates that fine-tuning LLMs with domain-specific data must be approached carefully, ensuring that foundational communication skills are preserved.

A comparison between open-source and proprietary models revealed significant differences in their performance profiles.
Open-source models often demonstrated ``spiky'' capabilities, excelling in specific dimensions while underperforming in others.
For instance, gpt-oss-120B achieved the highest $S_{\text{ACC}}$ and OpenBioLLM-70B led in $S_{\text{CES}}$, but both showed considerable deficits in other metrics.
In contrast, the proprietary models, GPT-4o and Claude Sonnet 4, exhibited a more balanced and consistently high performance across all CARE dimensions.
Although they did not always secure the highest score in every individual metric, their strong, well-rounded profiles suggest they are currently the most reliable choice for a comprehensive clinical conversational agent.
However, in scenarios where proprietary models are unfeasible due to data privacy constraints, data export limitations, or other restrictions, our findings indicate that HuatuoGPT-o1-72B presents itself as the most viable open-source alternative.

\section{Conclusion}
\label{sec:conclusion}
In this paper, we introduced AutoMedic, a novel, automated multi-agent framework that enables the evaluation of LLMs as clinical conversational agents through realistic, simulated dialogues.
To quantitatively assess performance within this framework, we also proposed the CARE metric, a multi-faceted tool that measures accuracy, conversational efficiency and strategy, empathy, and robustness.

Our investigation yielded several key findings.
Human expert evaluations confirmed the reliability of AutoMedic's patient profile generation and the strong correlation of our CARE metric with human clinical judgment.
We found that only a subset of medical QA datasets, such as MedBullets, MedQA, and MedXpertQA, are suitable for clinical simulation.
Crucially, we observed that all LLMs suffer a performance drop when moving from static QA to conversational settings.
This finding confirms that existing static medical QA benchmarks, which primarily assess knowledge, do not sufficiently reflect the actual performance of LLMs in dynamic medical field applications.
The CARE metric further revealed distinct LLM performance profiles.
Proprietary models like Claude Sonnet 4 and GPT-4o showed balanced, high-level performance, while most medical-tuned models underperformed, suggesting fine-tuning may erode communication skills.
HuatuoGPT-o1-72B, however, emerged as a strong open-source alternative.

This work provides a robust, automated methodology for the comprehensive evaluation of conversational medical AI, offering deeper insights than static metrics alone.
AutoMedic and the CARE metric can guide researchers and developers in building and selecting LLMs that are not only knowledgeable but also effective, empathetic, and reliable in practice.

\section*{Limitations}
While our framework provides a comprehensive evaluation of LLMs as clinical conversational agents, several limitations must be addressed.
First, our current evaluation is limited to text-based interactions and does not directly incorporate other modalities.
Given that many real-world medical scenarios are inherently multimodal and involve data such as medical imaging, a critical area for future work is extending the AutoMedic framework to evaluate vision-language and other multimodal models for broader impact.
Second, although the CARE metric provides a multi-faceted quantitative analysis, we did not define a single, aggregated score for overall performance.
A simple sum or average is insufficient, as the four dimensions have different scales and their relative importance can vary based on the specific clinical application.
Future research could address this by developing a weighted composite score, perhaps calibrated against human expert preferences derived from head-to-head model comparisons.
Finally, this paper focused exclusively on the evaluation of the doctor agent.
To ensure stability and consistency, we utilized a single proprietary LLM for the patient and clinical staff roles.
However, the performance and behavior of these supporting agents can influence the doctor agent's actions.
Analyzing the effect of different LLMs in these patient and staff roles on the overall simulation and doctor agent performance remains a valuable direction for future investigation.


\appendix

\section{Appendix}
\label{sec:appendix}

\begin{figure*}[!t]
    \centering
    \includegraphics[width=0.99\linewidth]{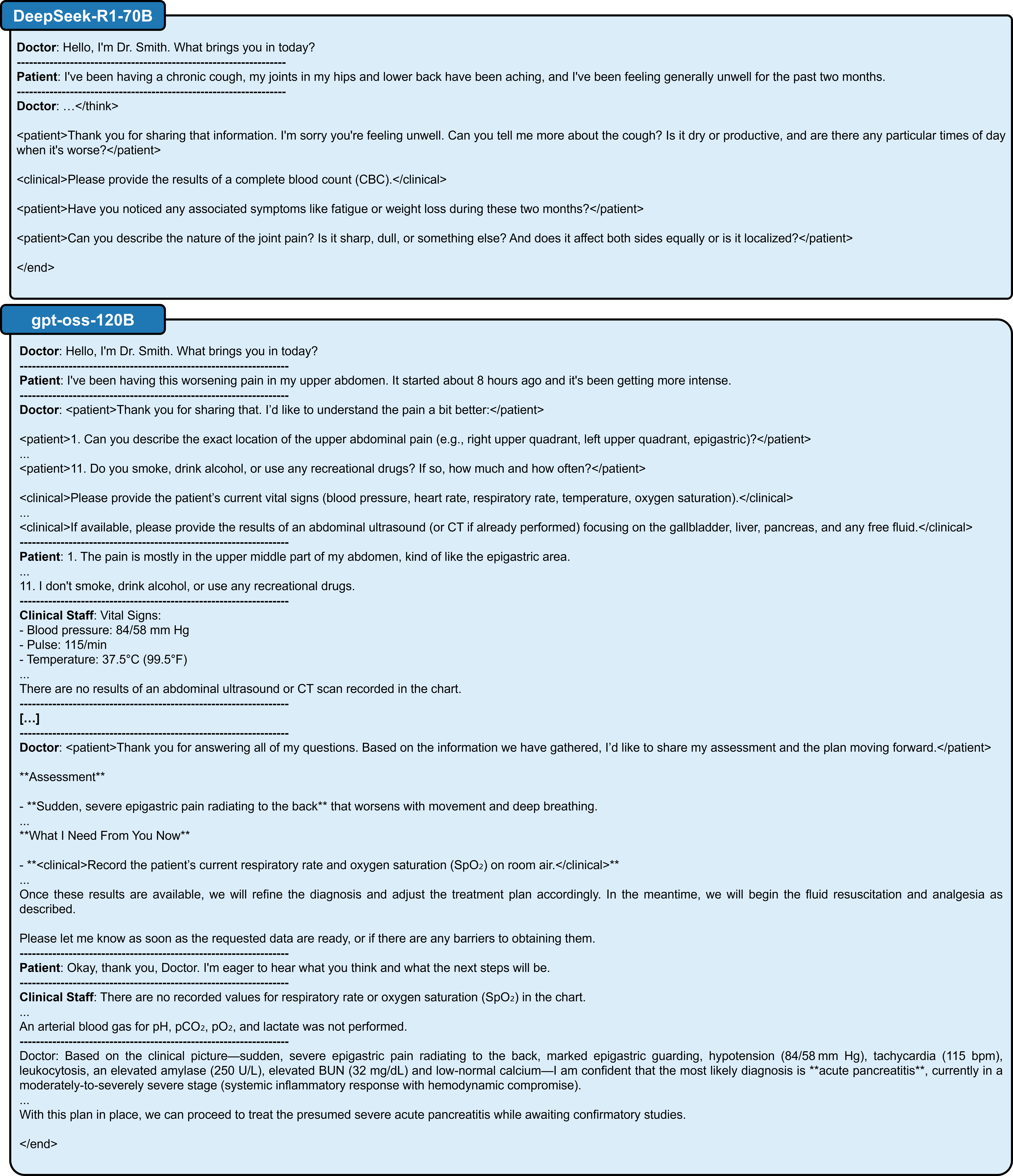}
    \caption{An example conversation generated by the DeepSeek-R1-70B and gpt-oss-120B doctor agent.}
    \label{fig:example_deepseek_gptoss}
\end{figure*}

\begin{figure*}[!ht]
    \centering
    \includegraphics[width=0.99\linewidth]{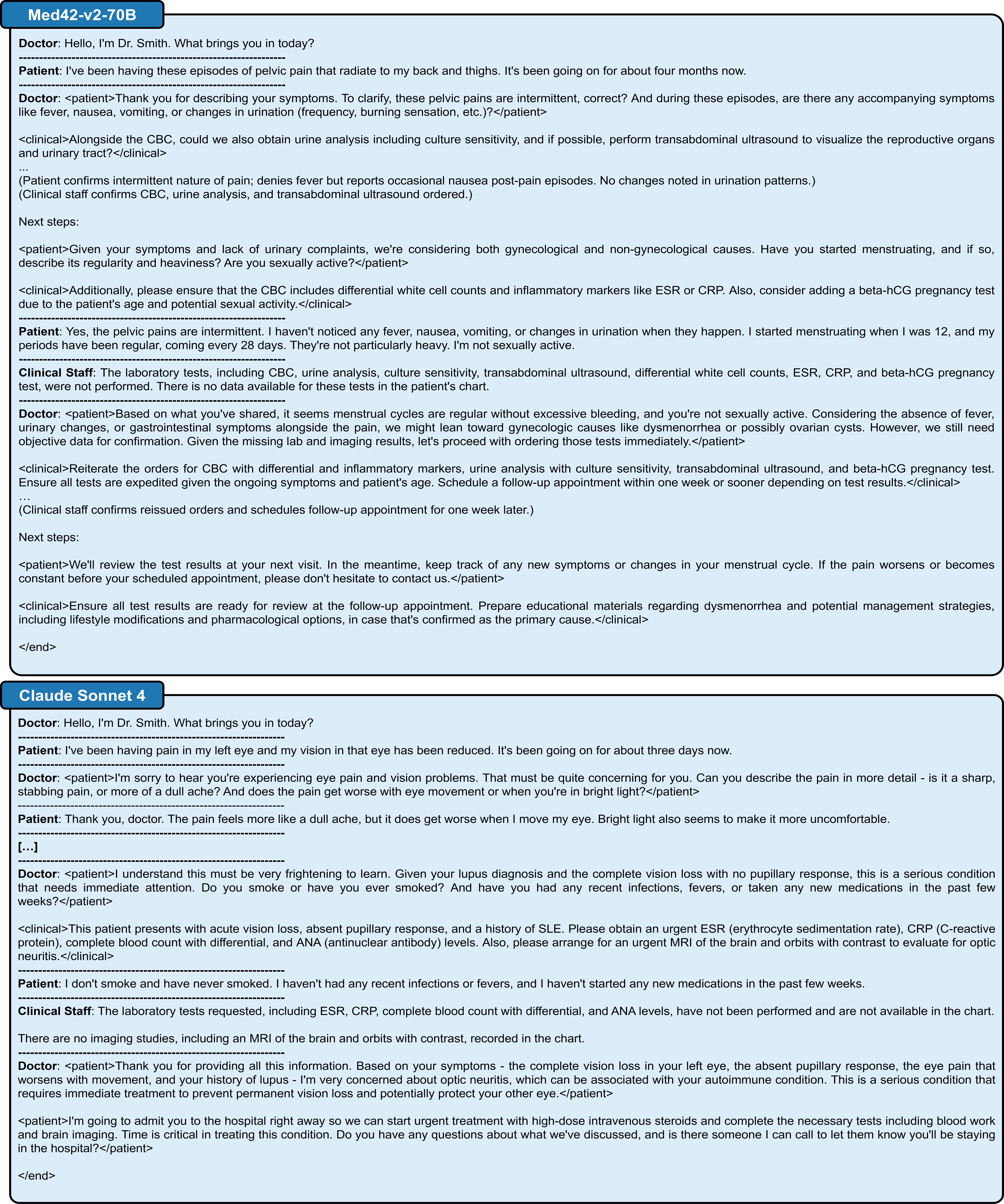}
    \caption{An example conversation generated by the Med42-v2-70B and Claude Sonnet 4 doctor agent.}
    \label{fig:example_med42_claude}
\end{figure*}

\end{document}